\documentclass[runningheads]{llncs}
\usepackage{graphicx}
\usepackage{amsmath,amssymb} 
\usepackage{color}

\usepackage{epsfig}
\usepackage{dsfont}
\usepackage{booktabs}
\usepackage{pifont}
\usepackage{multirow}
\usepackage[ruled]{algorithm2e}
\SetKwComment{Comment}{$\triangleright$\ }{}
\usepackage{subcaption}
\usepackage{bbold}

\newcommand{\etal}{\textit{et al}. }
\DeclareMathOperator*{\argmax}{arg\,max}

\begin{document}
\pagestyle{headings}
\mainmatter

\def\ACCV20SubNumber{759}  

\title{Double Targeted Universal Adversarial Perturbations} 
\titlerunning{Double Targeted UAPs}
%
\author{Philipp Benz$^*$ \and
Chaoning Zhang$^*$ \and
Tooba Imtiaz \and
In So Kweon}
\authorrunning{P. Benz et al.}
%
\institute{Korea Advanced Institute of Science and Technology (KAIST) \\
\email{pbenz@kaist.ac.kr, chaoningzhang1990@gmail.com}\\
$^*$ Equal contribution}

\maketitle

\begin{abstract}
Despite their impressive performance, deep neural networks (DNNs) are widely known to be vulnerable to adversarial attacks, which makes it challenging for them to be deployed in security-sensitive applications, such as autonomous driving. Image-dependent perturbations can fool a network for one specific image, while universal adversarial perturbations are capable of fooling a network for samples from all classes without selection. We introduce a double targeted universal adversarial perturbations (DT-UAPs) to bridge the gap between the instance-discriminative image-dependent perturbations and the generic universal perturbations. This universal perturbation attacks one targeted source class to sink class, while having a limited adversarial effect on other non-targeted source classes, for avoiding raising suspicions. Targeting the source and sink class simultaneously, we term it double targeted attack (DTA). This provides an attacker with the freedom to perform precise attacks on a DNN model while raising little suspicion. We show the effectiveness of the proposed DTA algorithm on a wide range of datasets and also demonstrate its potential as a physical attack. \footnote{Code: \url{https://github.com/phibenz/double-targeted-uap.pytorch}}
\end{abstract}

\section{Introduction}
Despite the recent success of deep learning~\cite{sutskever2012imagenet,hinton2012deep,collobert2008unified,zhang2020deepptz,zhang2019revisiting}, deep neural networks (DNNs) remain vulnerable to adversarial attacks~\cite{szegedy2013intriguing,goodfellow2014explaining,zhang2019cd-uap,zhang2020understanding,liu2019universal,benz2020data}. This poses a threat for deploying DNNs in security-sensitive applications, such as autonomous driving and robotics. Various attack methods~\cite{akhtar2018threat} have been proposed in the past few years, which can be roughly divided into two main categories: image-dependent attacks~\cite{szegedy2013intriguing,goodfellow2014explaining,kurakin2016adversarial,moosavi2016deepfool,carlini2017towards} and universal attacks~\cite{moosavi2017universal,khrulkov2018art,Mopuri2017datafree,metzen2017universal,poursaeed2018generative}. 
Image-dependent attacks construct perturbations tailored for a specific input image to be misclassified by the network; while universal attack methods aim to generate one single universal adversarial perturbation (UAP) that can fool the network for most samples of all classes. 
\begin{figure}
    \centering
    \includegraphics[width=0.7\textwidth]{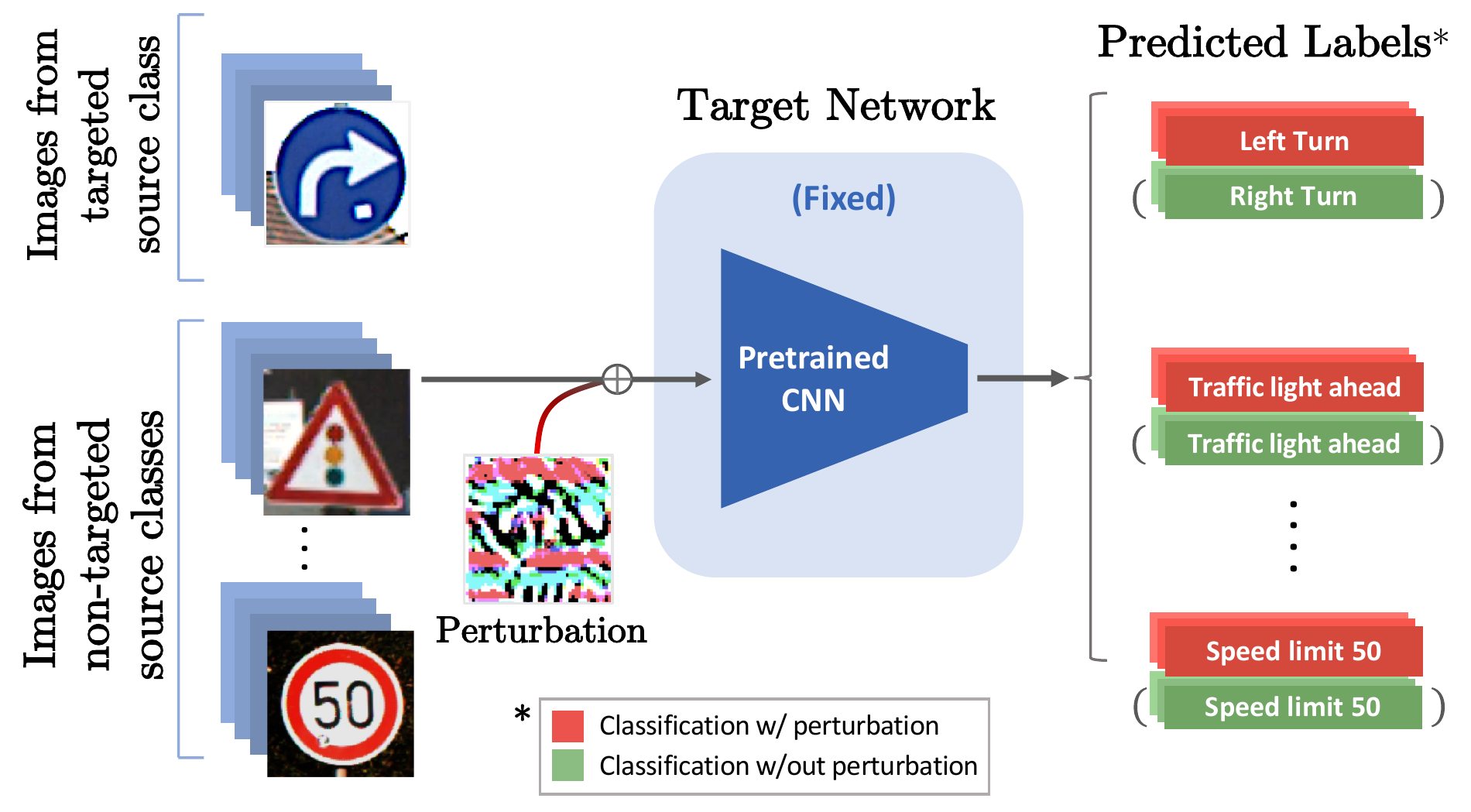}
    \caption{Overview of the Double Targeted Attack (DTA). In this example, the perturbation causes the network to classify images of the targeted source class \textbf{right turn} as the sink class \textbf{left turn}. Image classifications from the non-targeted source classes remain unaltered. The DT-UAP is added to all image samples.}
    \label{fig:teaser}
\end{figure}

Bridging the gap between the discriminative nature of image-dependent perturbations and the non-discriminative universal perturbation, we propose to attack a certain source class while limiting the influence of the attack on other, non-targeted classes. More specifically, we aim to fool the network with a single perturbation that can systematically shift a certain source class to a different sink class of choice. Since the proposed attack targets both the source and the sink class, we name it double targeted attack (DTA).
To avoid confusion, while other works~\cite{carlini2017towards,poursaeed2018generative} use the term ``target class'', we adopt ``sink class'' instead, since the proposed DTA also has target class(es) on the source side.

In this work, we focus on the exploration of universal perturbations due to their merit of being image-agnostic. This property eases the attack procedure for real-time applications such as autonomous driving or robotics, as the perturbation can be constructed in advance, and applying the prepared perturbation only requires one summation~\cite{moosavi2017universal}. UAPs attack all classes, making it obvious to an observer that a system is under attack. For achieving a more covert universal attack, class-discriminative universal adversarial perturbation (CD-UAP) has been introduced in~\cite{zhang2019cd-uap} to attack chosen class(es) on the source side. It would be more challenging yet meaningful to not only being class-discrimiantve on the source side, but also targets on the sink side. Compared with existing UAP attacks, DTA can be more dangerous in practice, since it allows precise attacks with flexible control over the targeted source class and the sink class. Applying double targeted universal adversarial perturbations (DT-UAP) can have fatal implications in practice. For instance, in the context of autonomous driving, an attacker can intentionally craft a perturbation to fool a network to misclassify traffic signs from ``turn left" to ``turn right" as shown in Fig.~\ref{fig:teaser}.

Technically, the proposed DTA does not strictly fall into the group of universal attacks, since it does not attack all classes. However, the DTA crafts one single perturbation that can be applied to the entire data distribution, which is similar to the existing UAPs~\cite{moosavi2017universal}. It is a non-trivial task to craft the DT-UAP because there is an inherent conflict between two objectives. For the samples from the targeted source class, the goal of the crafted perturbation is to shift their classification output to the sink class.
This will inevitably have a similar influence on the non-targeted source classes, which conflicts with the goal of the attack being discriminative between the targeted source class and other non-targeted source classes. Inspired by~\cite{zhang2019cd-uap}, we have designed an algorithm that explicitly deals with the trade-off between them.

To demonstrate its effectiveness, we evaluate the proposed DTA on five classification datasets from different domains for various DNN architectures. 

Our results establish the existence of DT-UAPs to attack data samples discriminatively. Though the designed DTA algorithm is mainly for perturbing samples to be misclassified from one targeted source class into one sink class, it can also be extended for shifting multiple targeted source classes to one sink class. We validate this specific attack scenario on the ImageNet dataset. Overall, our proposed algorithm has been validated to be effective to achieve discriminative targeted attacks with extensive experiments on different datasets and scenarios. Finally, we also demonstrate the potential of DTA being applied as a physical attack.

\section{Related work}

\subsection{Image-Dependent Attacks}
Adversarial attacks, which craft one perturbation specifically for one input image to fool a network are called image-dependent attacks.
Szegedy~\etal optimized such perturbations by using box-constrained L-BFGS~\cite{szegedy2013intriguing}.
Goodfellow~\etal then introduced the Fast Gradient Sign Method (FGSM), an efficient one-step attack to generate adversarial examples~\cite{goodfellow2014explaining}. The iterative variant of FGSM (I-FGSM) updates the perturbation by only a fraction of the allowed upper bound in each iteration~\cite{kurakin2016adversarial}.
Integrating the momentum term into the iterative process of I-FGSM (MI-FGSM) further improved the success rate of adversarial attacks ~\cite{dong2018boosting}.
DeepFool~\cite{moosavi2016deepfool} is also an iterative attack, manipulating the models' decision boundaries in the perturbation crafting process.
Incorporating the minimization of the perturbation magnitude into the optimization function, Carlini and Wagner (C\&W) introduced another three variants of image-dependent attacks~\cite{carlini2017towards}.
Another effective multi-step attack variant was introduced by Madry~\etal using projected gradient descent (PGD) to craft adversaries~\cite{madry2017towards}. The proposed DTA differentiates itself by attacking an entire class instead of only a single image.

\subsection{Universal Attacks}
A universal adversarial perturbation (UAP) is a single perturbation, which enables fooling a network for most input samples.
Accumulating image-dependent perturbations by iteratively applying DeepFool~\cite{moosavi2016deepfool}, Moosavi~\etal crafted the first UAPs~\cite{moosavi2017universal}.
In another variant, UAPs are crafted by leveraging the Jacobian matrices of the networks' hidden layers~\cite{khrulkov2018art}. Assuming no access to the original training data, Fast Feature Fool proposed to generate data-free UAPs by optimizing the feature change caused by the applied UAP~\cite{Mopuri2017datafree}. Generative Adversarial Perturbations (GAP) were proposed by Poursaeed~\etal~\cite{poursaeed2018generative}, using generative models to craft image-dependent and universal perturbations. Data-free targeted UAP has been introduced in~\cite{zhang2020understanding}, showing UAP have dominant features ovre images. The almost absent computational overhead (single summation) in the deployment of UAPs, makes them a favorable choice for the attack of real-world applications. Despite being universal, our proposed DTA differentiates itself from the existing universal attacks in its class-discriminative nature, i.e.\ by having a different influence on a sample depending on whether or not it belongs to the targeted source class. CD-UAP has been introduced in~\cite{zhang2019cd-uap}, our DT-UAP also targets on the sink side and thus constitutes a more challenging task. Moreover, we show that our DTA can also been used in physical attack~\cite{brown2017adversarial,liubias2020bias}.

\subsection{Attack on autonomous driving and robotics}
Deep learning has achieved the maturity to be deployed in safety and security-critical applications, such as autonomous driving~\cite{sallab2017deep} and robotics~\cite{sunderhauf2018limits}. The threat of adversarial attacks in these applications has also been widely explored. For example, Melis~\etal~\cite{melis2017deep} demonstrated the vulnerability of robots to the adversarially manipulated input images with the techniques in~\cite{szegedy2013intriguing}, and argue that secure robotics need to adopt strategies to enforce DNNs to learn more robust representations. Attack on the learning policy of robotics has been explored in~\cite{clark2018malicious}. Considering adversarial attacks in the context of autonomous driving, ~\cite{eykholt2018robust} generates UAPs to attack road sign classifiers. Another work~\cite{morgulis2019fooling} performs an attack in autonomous driving with traffic signs. Besides the classical classification dataset to evaluate the adversarial attack method, we also evaluate the proposed method on a traffic sign dataset and another robotics-related dataset.

\section{Double Targeted Attack}
\subsection{Problem Formulation}
The purpose of the proposed attack is to craft a single perturbation to shift one targeted source class to a different sink class. The source class to be attacked as well as the sink class are determined by the attacker to realize a flexible and precise attack. We term it double targeted attack (DTA).

Let $x \sim X$ denote a single sample from a distribution in $\mathds{R}^d$, and $\hat{F}(x)=p$ being a classification function, mapping input $x \in \mathds{R}^d$ to a predicted class $p \in [1,C]$ for a classification problem of $C$ classes. Here the classification function is represented through a DNN parameterized by the weights $\theta$.
For most samples from the targeted source class $x_t \sim X_t$, we seek a perturbation $\delta$ that satisfies the constraint
\begin{equation}
    \hat{F}(x_t+\delta) = y_{\text{sink}} \quad\text{subject to}\quad ||\delta ||_p \leq \epsilon,
    \label{eq1}
\end{equation}
where the sink class satisfies $y_{\text{sink}} \neq {F}(x_t)$, and $\epsilon$ indicates the magnitude limit for the $l_p$ norm of the crafted perturbation $\delta$. Note that limiting $X_t$ in Eq.~\ref{eq1} to a single image results in an image-dependent targeted attack. Meanwhile, it is equivalent to a non-discriminative targeted universal attack if the targeted samples $X_t$ comprise the entire dataset $X$.

Empirically, we find that a perturbation crafted under the constraint of Eq.~\ref{eq1} also shifts samples from the non-targeted source classes into the sink class with a high targeted fooling ratio. To incorporate covertness within the proposed attack, this effect of non-targeted samples $x_{nt} \sim X_{nt}$ shifting to the sink class should be minimized. The crafted perturbation should ideally shift instances from the chosen source class to a different sink class while having limited influence on the samples from the non-targeted source classes. More specifically, the proposed DTA has two objectives: 
(1) to increase the targeted fooling ratio for the samples from the chosen source class to the chosen sink class;
(2) to decrease the targeted fooling ratio for samples from the non-targeted source class(es) into the sink class, where the targeted fooling ratio is defined as the ratio of samples fooled into the sink class. These two objectives contradict each other, leading to an inevitable trade-off. In the following subsection, we state the loss function for DTA and design the algorithm for explicitly handling this trade-off between the two objectives. 

\subsection{DTA Loss design}

To achieve selectivity among the targeted source class and non-targeted source classes, we explicitly design different loss functions for the two. For the targeted class and the non-targeted classes, the loss is indicated by $\mathcal{L}_t$ and $\mathcal{L}_{nt}$, respectively. The final loss $\mathcal{L}$ can then be calculated as:
\begin{equation}
    \label{eq:loss}
    \mathcal{L} = \mathcal{L}_t + \alpha \mathcal{L}_{nt},
\end{equation}
where $\alpha$ is a hyper-parameter for weighting the trade-off between $\mathcal{L}_{t}$ and $\mathcal{L}_{nt}$. In practice, this hyper-parameter can be fine-tuned by the attacker for a specific task. For simplicity, we set $\alpha$ to $1$ in all of our experiments. We empirically found that this setting works well when the same number of samples are sampled from $X_t$ and $X_{nt}$ in every iteration update. 

For the targeted class, the loss $\mathcal{L}_{t}$ should shape the perturbation to fool the network by shifting the prediction from the source class into the sink class. This can be realized through (1) decreasing the logit value for the originally predicted class $\hat{L}_p$ with $p = \argmax(\hat{L}(x_t))$ to not being the highest logit anymore, while (2) increasing the logit for the sink class $\hat{L}_{\text{sink}}$, to be the dominant logit, where $\hat{L}(\cdot)$ indicates the function mapping to the logit values and $\hat{L}_i$ is the specific logit value of class $i$. Thus, $\mathcal{L}_{t}$ can be decomposed into two parts as follows:
\begin{gather}
    \mathcal{L}_{t} = \mathcal{L}_{t1} + \mathcal{L}_{t2} \text{, with} \\
    \mathcal{L}_{t1} = \max(\hat{L}_p(x_t + \delta) - \underset{i \neq p}\max(\hat{L}_i(x_t + \delta)), 0) \\ 
    \mathcal{L}_{t2} = \max(\underset{i \neq y_{\text{sink}}}\max(\hat{L}_i(x_t + \delta) - \hat{L}_{\text{sink}}(x_t + \delta)), -D)
\end{gather}
where the hyper-parameter $D$ constitutes an intensity value of the dominance of the targeted logit value. A higher $D$ implies a higher chance that the sample will be classified as the sink class. For the non-targeted source classes, we adopt the widely used cross-entropy function as:
\begin{equation}
    \mathcal{L}_{nt} = \mathcal{X}(\hat{L}(x_{nt} + \delta), \mathbb{1}(\hat{F}(x_{nt})))
\end{equation}
with $\mathbb{1}(\cdot)$ indicating a one-hot encoded vector of $C$ classes. In practice, an attacker can change the hyper-parameters according to the requirements. For instance, the attacker can increase the parameter $\alpha$ in Eq.~\ref{eq:loss} to increase the covertness of the proposed attack accompanied by a relatively low targeted fooling ratio for the targeted class, or increase the parameter $D$ in order to achieve stronger classifications into the sink class.

To balance the two contradicting objectives, clamping of the logit values was adopted in $\mathcal{L}_{t}$. Without this clamping operation, the loss part of the targeted classes $\mathcal{L}_t$ can prevail by shifting the samples from the targeted source class to the sink class, while disregarding the other objective of limiting the influence on samples from the non-targeted classes. Since this loss clamping is applied to every targeted source class sample in the batch, it can also facilitate avoiding any sample dominating over other samples for contributing to the gradient of the universal perturbation. A similar clamping technique has been applied in~\cite{carlini2017towards} but with the objective to achieve a minimum-magnitude (image-dependent) perturbation that can attack a specific sample.

\subsection{DTA Algorithm}
\begin{algorithm}[t]
\small
    \SetAlgoLined
    \DontPrintSemicolon
    \SetKwInput{KwInput}{Input}
    \SetKwInput{KwOutput}{Output}
    \SetKwFunction{FOptim}{Optim}
    \KwInput{Data distribution $X$, Classifier $\hat{F}$, Loss function $\mathcal{L}$, Mini-batch size $m$, Number of iterations $I$, Perturbation magnitude $\epsilon$}
    \KwOutput{Perturbation vector $\delta$}
    $X_t \subseteq X$ \Comment*[r]{Subset}
    $X_{nt} \subseteq X$ \Comment*[r]{Subset}
    $\delta \leftarrow 0$ \Comment*[r]{Initialize}
    \For {iteration $=1, \dots, I$}{
        $B_t \sim X_t$: $|B_t| =  \frac{m}{2}$ \Comment*[r]{Randomly sample}
        $B_{nt} \sim X_{nt}$: $|B_{nt}| = \frac{m}{2}$ \Comment*[r]{Randomly sample}
        $B \leftarrow B_t \bigcup B_{nt}$ \Comment*[r]{Concatenate} 
        $g_{\delta} \leftarrow \underset{B}{\mathds{E}} [\nabla_{\delta} \mathcal{L}$] \Comment*[r]{Calculate gradient} 
        $\delta \leftarrow$ \FOptim{$g_\delta$} \Comment*[r]{Update perturbation} 
        $\delta \leftarrow \frac{\delta}{||\delta||_p} \epsilon$ \Comment*[r]{Projection} 
        }
\caption{Double Targeted Attack Algorithm}
\label{alg:dta}
\end{algorithm}

With the loss functions defined above, the procedure to craft DT-UAPs with DTA is shown in Algorithm~\ref{alg:dta}. For each perturbation update iteration, we include samples from both the targeted source class and the non-targeted source classes. More specifically, we randomly select the same number (half of the mini-batch size) of samples from the targeted source class and the non-targeted source classes to form $B_t$ and $B_{nt}$, which can be concatenated to one batch $B$. We then calculate the loss parts $\mathcal{L}_{t}$ and $\mathcal{L}_{nt}$ referring to Eq. $3$ and Eq. $6$, respectively. The total loss $\mathcal{L}$ can then be calculated referring to Eq.~\ref{eq:loss}. This procedure illustrates how the loss $\mathcal{L}$ in Algorithm~\ref{alg:dta} is calculated. 
The perturbation can then be updated with the loss gradient calculated with respect to the perturbation. 
Note that the gradient thus computed is the expected gradient, i.e.\ the average of the gradients in this mini-batch. For the update of the perturbation, we can adopt any existing optimizer, but we empirically found that the ADAM~\cite{kingma2014adam} optimizer converges the fastest for our method. In the final step, the perturbation is projected to the $l_p$-ball with radius $\epsilon$ in order to satisfy the magnitude constraint. This process is repeated for $I$ iterations. Mini-batch training and balancing the sample amount from the two data distributions result in a simple yet effective algorithm. Our algorithm is mainly inspired by~\cite{zhang2019cd-uap,zhang2020understanding}. Their algorithm has been shown to outperform UAP~\cite{moosavi2017universal} and GAP~\cite{poursaeed2018generative} by a large margin, achieving SOTA performance for universal attack. Here, we tailor it to suite our purpose of being double targeted.

\section{Results and Analysis}
\subsection{Experimental Setup}

We apply the proposed DTA to various deep convolutional neural network architectures and construct perturbations on various datasets: CIFAR-10 \cite{Krizhevsky09CIFAR}, GTSRB~\cite{stallkamp2012man}, EuroSAT~\cite{helber2019eurosat}, YCB~\cite{calli2015ycb} and large-scale ImageNet~\cite{deng2009imagenet}. CIFAR-10 and ImageNet are two commonly used benchmark datasets for image classification tasks. The GTSRB dataset consists of $43$ classes of different German traffic signs and is a commonly used dataset for autonomous driving applications. The EuroSAT dataset is used for land cover classification tasks via satellite images categorized into $10$ classes. The YCB dataset is a benchmark dataset for robotic manipulation and consists of a total of $98$ classes of daily life objects. 

For the different datasets, we evaluate DTA with at least two different networks. Overall, we explore various DNN architectures, including VGG-16~\cite{simonyan2014very}, ResNet-20/50~\cite{he2016identity}, Inception-V3~\cite{szegedy2016rethinking} and MobileNet-V2~\cite{sandler2018mobilenetv2}. To evaluate our approach, we use the metric of the targeted fooling ratio $\kappa$, which is defined as the ratio of samples fooled into the sink class. We apply the targeted fooling ratio to the targeted source class and non-targeted source classes, indicated by $\kappa_t$ and $\kappa_{nt}$, respectively. Consequently, the higher (lower) $\kappa_t$ ($\kappa_{nt}$), the better. For the following experiments, we set the number of iterations to $I = 500$, adopt the $l_{\infty}$ norm and cap the perturbation magnitude at $\epsilon=15$ for images in the range $[0,255]$. All our experiments are performed using the PyTorch (v.$0.4.1$)~\cite{paszke2019pytorch} framework on a single GPU TITAN X (Pascal). Note that for crafting the perturbation, we only use the correctly classified images from the training dataset and report the results on all samples from the validation dataset.

\subsection{Quantitative Results}

\begin{table}[t]
\centering
\caption{Experimental results for the Double Targeted Attack (DTA) for the datasets CIFAR-10, GTSRB, EuroSAT, YCB and ImageNet under $10$ scenarios $S_0$ to $S_9$. For each scenario, the targeted fooling ratios for the targeted source samples ($\kappa_t$) and the non-targeted source samples ($\kappa_{nt}$) are reported. All numbers are reported in $\%$.}
\label{tab:results}
\small
\setlength\tabcolsep{1.5pt}
\scalebox{0.65}{
\begin{tabular}{cc |cc|cc|cc|cc|cc|cc|cc|cc|cc|cc|cc}
\toprule
\multirow{2}{*}{Dataset} & \multirow{2}{*}{Model} & \multicolumn{2}{c}{$S_0$} & \multicolumn{2}{c}{$S_1$} & \multicolumn{2}{c}{$S_2$} & \multicolumn{2}{c}{$S_3$} & \multicolumn{2}{c}{$S_4$} & \multicolumn{2}{c}{$S_5$} & \multicolumn{2}{c}{$S_6$} & \multicolumn{2}{c}{$S_7$} & \multicolumn{2}{c}{$S_8$} & \multicolumn{2}{c}{$S_9$} & \multicolumn{2}{c}{Avg} \\
& & $\kappa_t$ &  $\kappa_{nt}$ & $\kappa_t$ &  $\kappa_{nt}$ & $\kappa_t$ &  $\kappa_{nt}$ & $\kappa_t$ &  $\kappa_{nt}$ & $\kappa_t$ &  $\kappa_{nt}$ & $\kappa_t$ &  $\kappa_{nt}$ & $\kappa_t$ &  $\kappa_{nt}$ & $\kappa_t$ &  $\kappa_{nt}$ & $\kappa_t$ &  $\kappa_{nt}$ & $\kappa_t$ &  $\kappa_{nt}$ & $\kappa_t$ &  $\kappa_{nt}$ \\
\midrule
\multirow{2}{*}{CIFAR-10} & VGG-16       & $77.5$ & $20.5$ & $83.5$ & $22.0$ & $78.2$ & $14.7$ & $81.4$ & $21.5$ & $73.0$ & $18.6$ & $79.1$ & $14.2$ & $75.1$ & $15.1$ & $76.7$ & $24.6$ & $75.0$ & $20.3$ & $86.2$ & $16.6$ & $78.6$ & $18.8$  \\
                          & ResNet-20    & $78.8$ & $26.1$ & $84.6$ & $28.0$ & $84.0$ & $24.3$ & $84.2$ & $26.9$ & $77.1$ & $22.0$ & $82.1$ & $21.3$ & $83.8$ & $14.7$ & $72.9$ & $33.2$ & $80.0$ & $27.8$ & $89.8$ & $22.3$ & $81.7$ & $24.7$ \\ 
\midrule
\multirow{2}{*}{GTSRB}    & VGG-16       & $89.0$ & $0.2$  & $100$ & $1.1$  & $87.1$ & $1.2$  & $72.2$ & $0.6$  & $91.0$ & $1.3$  & $83.6$ & $2.4$  & $88.3$ & $1.1$  & $80.0$ & $0.7$  & $95.0$ & $1.9$  & $81.1$  & $1.7$  & $86.7$ & $1.2$ \\
                          & ResNet-20    & $84.3$ & $0.5$  & $100$ & $1.6$  & $53.1$ & $0.2$  & $77.8$ & $1.8$  & $87.6$ & $2.9$  & $77.1$ & $4.4$  & $70.0$ & $2.7$  & $88.3$ & $1.2$  & $80.0$ & $0.3$  & $64.4$  & $0.7$ & $78.3$ & $1.6$ \\
\midrule
\multirow{2}{*}{EuroSAT}  & ResNet-50    & $96.2$ & $33.0$ & $98.8$ & $18.0$ & $95.2$ & $31.1$ & $96.6$ & $22.1$ & $99.2$ & $28.7$ & $95.0$ & $24.0$ & $94.4$ & $44.3$ & $96.3$ & $17.6$ & $96.3$ & $24.5$ & $91.2$ & $22.7$ & $95.9$ & $26.6$ \\
                          & Inception-V3 & $94.3$ & $28.7$ & $95.2$ & $18.9$ & $93.8$ & $41.4$ & $99.2$ & $56.3$ & $93.0$ & $29.4$ & $93.0$ & $24.2$ & $91.6$ & $34.6$ & $96.0$ & $21.8$ & $96.8$ & $31.6$ & $89.2$ & $18.8$ & $94.2$ & $30.6$ \\ 
\midrule 
\multirow{2}{*}{YCB}      
                          & ResNet-50    & $100$ & $14.5$ & $100$ & $24.2$ & $100$ & $32.4$ & $96.7$ & $38.0$ & $100$ & $33.5$  & $99.2$ & $38.3$ & $100$ & $44.4$ & $99.2$ & $41.7$ & $100$ & $19.0$ & $100$ & $33.1$ &$99.5$ & $31.9$ \\
                          & Inception-V3 & $100$ & $16.6$ & $100$ & $30.0$ & $100$ & $38.7$ & $99.2$ & $31.2$ & $100$ & $12.9$ & $98.3$ & $20.0$ & $100$ & $32.2$ & $100$ & $36.6$ & $100$ & $17.3$ & $100$ & $39.2$ & $99.8$ & $27.5$ \\

\midrule 
\multirow{4}{*}{ImageNet} & VGG-16       & $72.0$ & $10.3$ & $96.0$ & $19.5$ & $90.0$ & $19.5$ & $82.0$ & $28.3$ & $74.0$ & $15.9$ & $82.0$ & $13.0$ & $66.0$ & $8.9$  & $64.0$ & $12.9$ & $66.0$ & $21.5$ & $70.0$ & $26.1$ & $76.2$ & $17.6$ \\
                          & ResNet-50    & $74.0$ & $13.9$ & $94.0$ & $21.4$ & $82.0$ & $15.2$ & $72.0$ & $20.9$ & $62.0$ & $13.6$ & $84.0$ & $15.5$ & $72.0$ & $9.8$  & $66.0$ & $21.4$ & $66.0$ & $17.3$ & $62.0$ & $18.1$ & $73.4$ & $16.7$ \\
                          & Inception-V3 & $78.0$ & $10.0$ & $86.0$ & $15.7$ & $86.0$ & $12.2$ & $78.0$ & $15.6$ & $58.0$ & $9.5$  & $76.0$ & $12.9$ & $70.0$ & $8.9$  & $72.0$ & $15.7$ & $62.0$ & $18.9$ & $66.0$ & $17.8$ & $73.2$ & $13.7$ \\
                          & MobileNet-V2 & $74.0$ & $11.3$ & $94.0$ & $17.0$ & $88.0$ & $20.4$ & $70.0$ & $15.3$ & $72.0$ & $16.0$ & $84.0$ & $15.0$ & $74.0$ & $14.5$ & $74.0$ & $21.7$ & $72.0$ & $18.8$ & $70.0$ & $21.9$ & $77.2$ & $17.2$ \\ 
\bottomrule
\end{tabular}
}
\end{table}

\begin{table*}[t]
\centering
\caption{Targeted source class to sink class mapping for the datasets CIFAR-10, GTSRB, YCB, EuroSAT, and ImageNet.}
\label{tab:scenarios1}
\setlength\tabcolsep{1.5pt}
\scalebox{0.5}{
\begin{tabular}{cccccc}
\toprule
 $S$ & CIFAR-10 & GTSRB & YCB & EuroSAT & ImageNet \\
\midrule
$S_0$  & bird $\rightarrow$ airplane       & turn right ahead $\rightarrow$ turn left ahead       & large clamp $\rightarrow$ strawberry             & Herb. Vegetation $\rightarrow$ Annual Crop & wig $\rightarrow$ lab coat            \\
$S_1$  & deer $\rightarrow$ frog           & end prev. limitation $\rightarrow$ end no passing    & flat screwdriver $\rightarrow$  mini soccer ball & Industrial $\rightarrow$ Permanent Crop    & photocopier $\rightarrow$ castle      \\
$S_2$  & frog $\rightarrow$ cat            & no passing $\rightarrow$ no Lkw permitted            & cups type f $\rightarrow$ larger marker          & Permanent Crop $\rightarrow$ Highway       & flagpole $\rightarrow$ sewing machine \\
$S_3$  & ship $\rightarrow$ cat            & wild animals possible $\rightarrow$ bicycle lane     & hammer $\rightarrow$ lego duplo type i           & River $\rightarrow$ Highway                & jersey $\rightarrow$ rain barrel      \\
$S_4$  & truck $\rightarrow$ horse         & no vehicles permitted $\rightarrow$ speed limit $70$ & cups type c $\rightarrow$ toy airplane part i    & Sea Lake $\rightarrow$ Residential         & theater curtain $\rightarrow$ brass   \\
$S_5$  & airplane $\rightarrow$ deer       & no passing $\rightarrow$ speed limit $60$            & tuna fish can $\rightarrow$ plastic nut          & Residential $\rightarrow$ Pasture          &  drilling platf. $\rightarrow$ pomegranate \\
$S_6$  & horse $\rightarrow$ dog           & slippery road $\rightarrow$ uneven surfaces          & tomato soup can $\rightarrow$ cups type g        & Permanent Crop $\rightarrow$ River         &  fireboat $\rightarrow$ aircraft carrier \\
$S_7$  & dog $\rightarrow$ frog            & pedestrian crossing $\rightarrow$ double curves      & cups type h $\rightarrow$ chain                  & Pasture $\rightarrow$ Permanent Crop       &  torch $\rightarrow$ golfcart \\
$S_8$  & dog $\rightarrow$ deer            & speed limit $20$ $\rightarrow$ Speed Limit $120$     & marbles type $3$ $\rightarrow$ key               & Pasture $\rightarrow$ Industrial           &  candle $\rightarrow$ howler monkey \\
$S_9$ & airplane $\rightarrow$ automobile  & road narrows right $\rightarrow$ children crossing   & cups type j $\rightarrow$ toy airplane part k    & Annual Crop $\rightarrow$ Forest           &  ruddy turnstone $\rightarrow$ kuvasz \\
\bottomrule
\end{tabular}
}
\end{table*}

We evaluate the effectiveness of the proposed DTA by randomly selecting $10$ source-to-sink shift scenarios indicated by $S_0$ to $S_9$ for each dataset. The results are summarized in Table~\ref{tab:results}, where we report the targeted fooling ratio for both the targeted class $\kappa_t$ and the non-targeted classes $\kappa_{nt}$ for each scenario. The exact mapping of the targeted source class to the sink class can be found in Table~\ref{tab:scenarios1}.

Overall, the results in Table~\ref{tab:results} indicate that DTA achieves reasonable performance for different mapping scenarios on a wide range of datasets. This conclusion stems from two major observations. First, the targeted fooling ratio for the targeted classes ($\kappa_t$) is quite high. Second, there is a significant gap between $\kappa_{t}$ and $\kappa_{nt}$, which indicates that the crafted perturbation is discriminative between targeted class and non-targeted classes.
We further analyze the performance of each dataset.

\subsubsection{CIFAR-10}
With an average $\kappa_t$ of around $80\%$, DTA performs reasonably well on CIFAR-10, fooling most of the targeted source class into the sink class. The gap between $\kappa_t$ and $\kappa_{nt}$ is about $58\%$, indicating sufficient selectivity.

\subsubsection{GTSRB}
For the task of road sign classification, our proposed DTA can even achieve a $100\%$ targeted fooling ratio for scenario $S_1$ while maintaining a very low targeted fooling ratio of $1.1\%$ and $1.6\%$ for VGG-16 and ResNet-20, respectively, on the non-targeted source samples. Overall, DTA exhibits high $\kappa_t$ values, while maintaining the lowest $\kappa_{nt}$ values among all examined datasets. Therefore, DTA achieves the highest gap between $\kappa_{t}$ and $\kappa_{nt}$ for the GTSRB dataset. The low $\kappa_{nt}$ indicates that the perturbations for attacking GTSRB are especially covert. We speculate that the reason behind the high performance on the GTSRB dataset is that the in-class variation is very small, making the discriminative attack a relatively easy task.

\subsubsection{EuroSAT and YCB}
The results of DTA on the EuroSAT and YCB datasets exhibit similar behavior, with very high values for $\kappa_t$ ,above $94\%$, while having $\kappa_{nt}$ values of around $30\%$. With a gap of more than $60\%$, DTA poses a strong, covert threat for applications deploying satellite images and classification tasks for robotic manipulation. 

\subsubsection{ImageNet}
The results show that DTA is able to fool a network for a single class out of the $1000$ into a sink class for all $4$ investigated DNNs, namely VGG-16, ResNet-50, Inception-V3, and MobileNet-V2. For specific scenarios such as $S_3$ or $S_4$, there can be a relatively large performance gap among different DNN architectures. Overall, with an average $\kappa_t$ of around $75\%$ and an average $\kappa_{nt}$ of $16\%$, different DNNs have comparable performance.

\subsection{Qualitative Results}

\begin{figure}[t]
    \centering
    \includegraphics[width=0.19\linewidth]{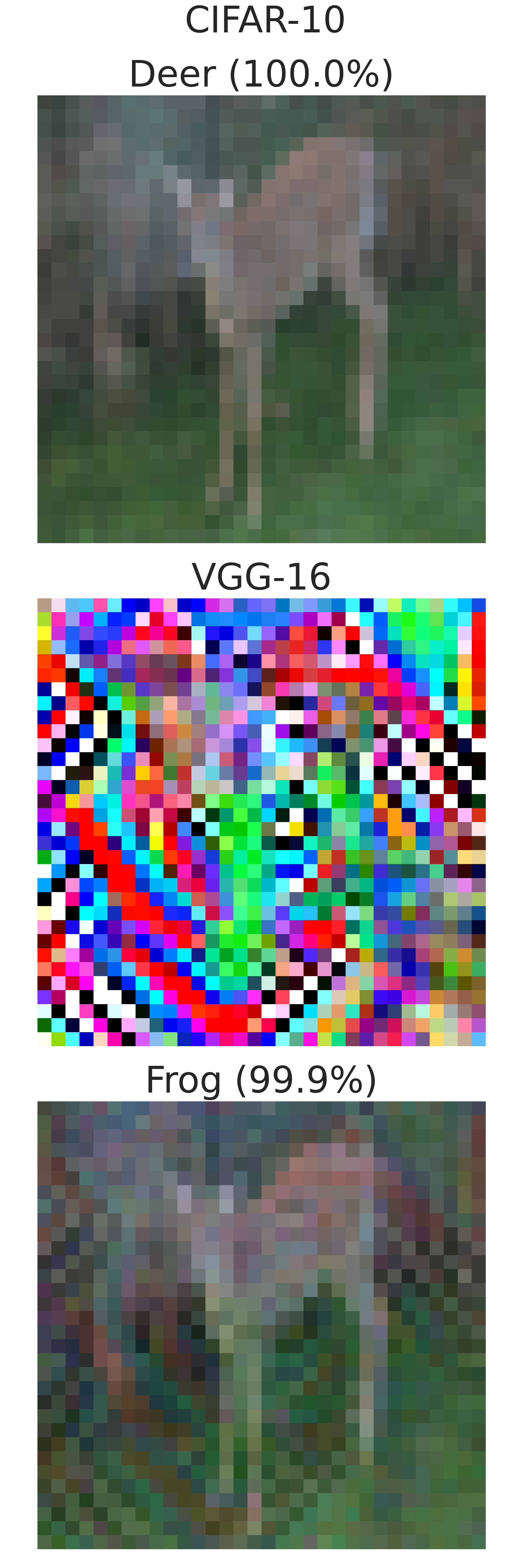}
    \includegraphics[width=0.19\linewidth]{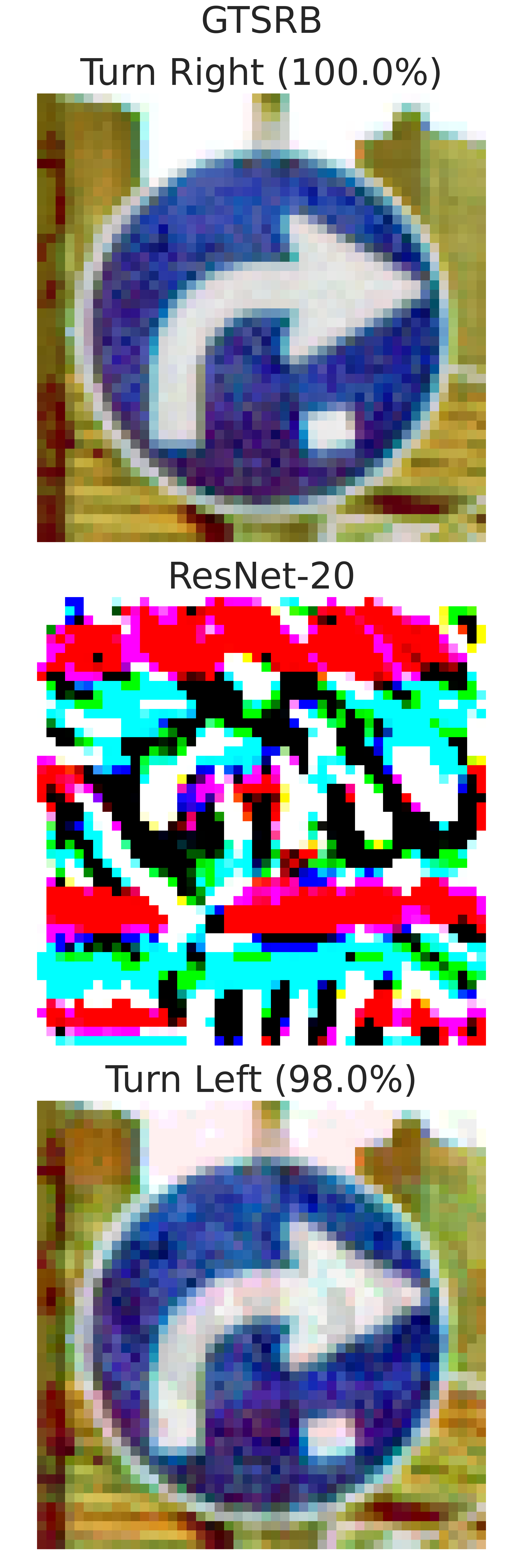}
    \includegraphics[width=0.19\linewidth]{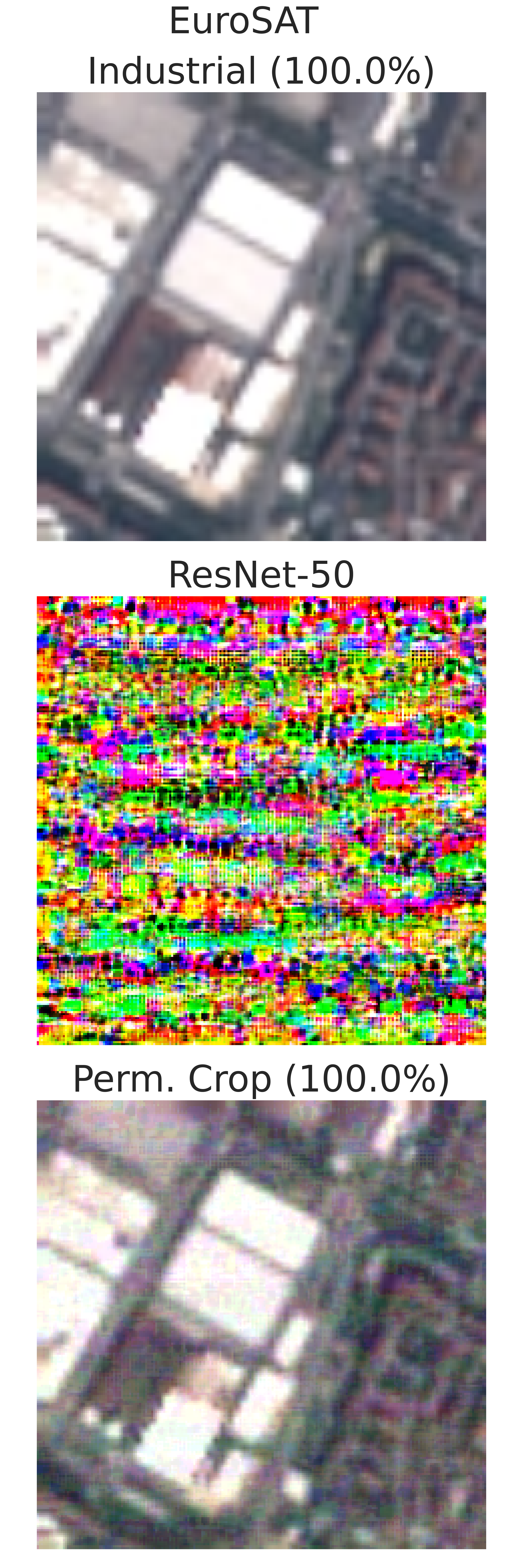}
    \includegraphics[width=0.19\linewidth]{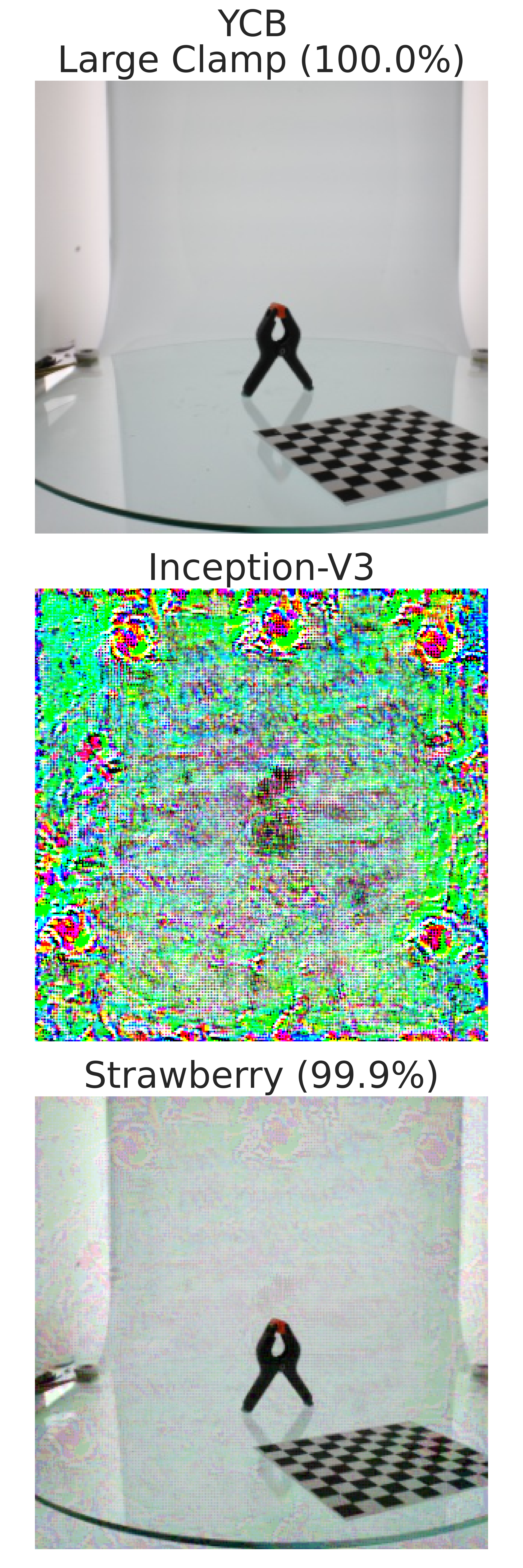}
    \includegraphics[width=0.19\linewidth]{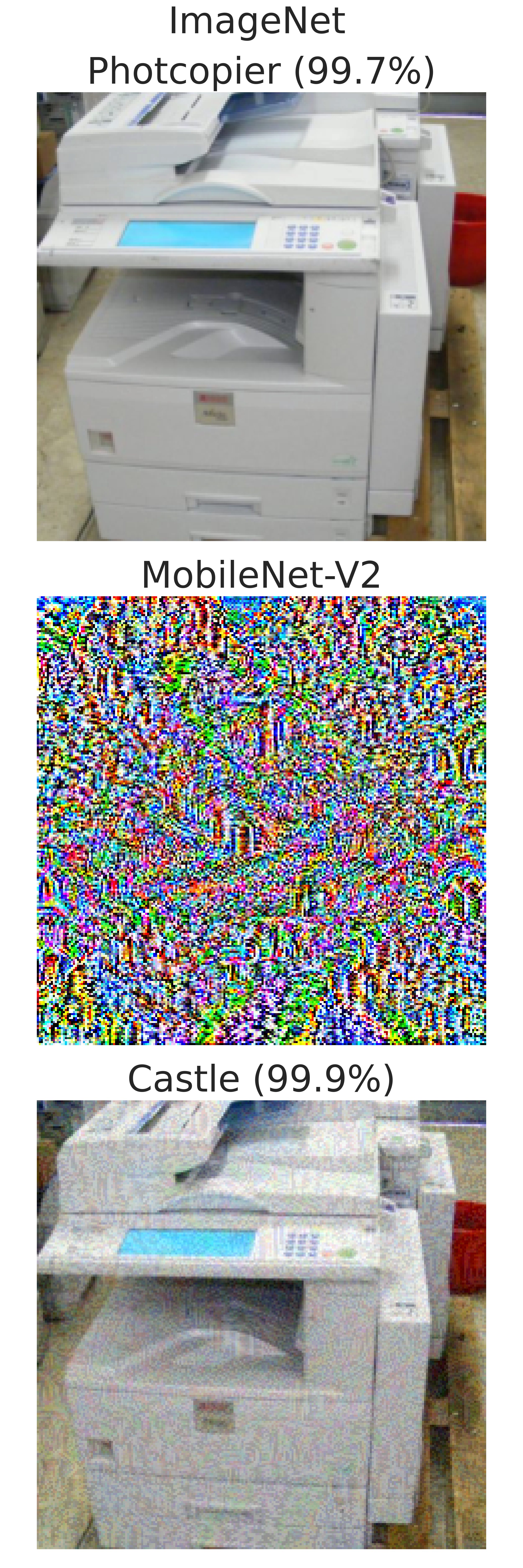}
    \caption{Examples of adversarial perturbations for various datasets and networks. The figure shows the original images (top), an amplified version of the corresponding perturbations (middle) and the resulting adversarial examples (bottom). The confidence values of the network and the predicted labels are stated above the images. The target network is indicated above the amplified perturbation.}
    \label{fig:qual_res}
\end{figure}

In this subsection, we illustrate perturbations and perturbed samples generated by the proposed DTA. Fig.~\ref{fig:qual_res} shows the original targeted source image, along with the amplified universal perturbation and the resulting adversarial image. It can be observed that the DTA produces patterns with different characteristics for each dataset. The adversarial image is still identifiable as a source class instance to a human observer, however, the DNN classifies the manipulated image (from the targeted source class) with high confidence into the sink class. 

\subsection{Universal Multi2One Targeted Perturbation}
Finally, we extend the DT-UAPs to a more challenging scenario to demonstrate an extension of the DTA. To this end, we alter the objective from one targeted source class to instead support multiple source classes ($MS$) while still leading the samples from these classes to one sink class. Due to this property of classifying multiple source classes to one sink class, we term the resulting perturbation a universal Multi2One targeted perturbation. Crafting such perturbations is more challenging since multiple source classes add complexity which has to be compensated by the universal perturbation. We evaluate this attack for $4$ scenarios, which are detailed in Table~\ref{tab:classes_multi2one} under the same settings as before. The results in Table~\ref{tab:results_multi2one} show that our proposed DTA also achieves reasonable performance in the case of shifting multiple targeted source classes into the sink class.

\begin{table}[t]
\centering
\caption{Experimental results for the universal Multi2One targeted perturbation on ImageNet under $4$ scenarios $MS_0$ to $MS_3$. For each scenario, $\kappa_t$ $\kappa_{nt}$ are reported. All numbers are reported in $\%$.}
\label{tab:results_multi2one}
\small
\setlength\tabcolsep{1.5pt}
\scalebox{0.95}{
\begin{tabular}{c |cc|cc|cc|cc|cc}
\toprule
\multirow{2}{*}{Model} & \multicolumn{2}{c}{$MS_0$} & \multicolumn{2}{c}{$MS_1$} & \multicolumn{2}{c}{$MS_2$} & \multicolumn{2}{c}{$MS_3$} & \multicolumn{2}{c}{Avg} \\
& $\kappa_t$ &  $\kappa_{nt}$ & $\kappa_t$ &  $\kappa_{nt}$ & $\kappa_t$ &  $\kappa_{nt}$ & $\kappa_t$ &  $\kappa_{nt}$ & $\kappa_t$ &  $\kappa_{nt}$ \\
\midrule
VGG16        & $63.3$ & $24.9$ & $69.3$ & $33.7$ & $76.0$ & $25.8$ & $69.3$ & $26.8$ & $69.5$ & $27.8$ \\
ResNet-50    & $64.0$ & $30.1$ & $63.3$ & $32.3$ & $78.7$ & $29.2$ & $62.7$ & $23.2$ & $67.2$ & $28.7$ \\
Inception-V3 & $58.0$ & $19.4$ & $56.7$ & $23.8$ & $66.7$ & $19.0$ & $66.7$ & $20.8$ & $62.0$ & $20.8$ \\
MobileNet-V2 & $68.0$ & $27.2$ & $66.0$ & $28.0$ & $74.0$ & $25.6$ & $66.0$ & $24.4$ & $68.5$ & $26.3$ \\
\bottomrule
\end{tabular}
}
\end{table}

\begin{table}[t]
\centering
\caption{Targeted source classes to sink class mapping for the Multi2One attack on ImageNet.}
\label{tab:classes_multi2one}
\small
\setlength\tabcolsep{2.pt}
\scalebox{0.95}{
\begin{tabular}{ccccc}
 $MS$ & ImageNet  \\\toprule
$MS_0$  & affenpinscher, black grouse, alp $\rightarrow$ mosque \\
$MS_1$  & necklace, four-poster, jersey  $\rightarrow$ llama \\
$MS_2$  & wig, photocopier, flagpole $\rightarrow$ castle \\
$MS_3$  & granny smith, dragonfly, drilling platform $\rightarrow$ brass\\
\bottomrule
\end{tabular}
}
\end{table}

\subsection{Ablation Analysis}

\begin{table}[t]
\centering
\caption{Analysis of the influence of different loss function configurations. For each scenario of the 4 different scenarios, $\kappa_t$ and $\kappa_{nt}$ are reported. All numbers are reported in $\%$.}
\label{tab:results_no_sink_loss}
\small
\setlength\tabcolsep{2.0pt}
\scalebox{0.95}{
\begin{tabular}{c| c |cc|cc|cc|cc|cc}
\toprule
\multirow{2}{*}{$\mathcal{L}$} & \multirow{2}{*}{Dataset} & \multicolumn{2}{c}{$S_0$} & \multicolumn{2}{c}{$S_1$} & \multicolumn{2}{c}{$S_2$} & \multicolumn{2}{c}{$S_3$} & \multicolumn{2}{c}{Avg}  \\
& & $\kappa_t$ &  $\kappa_{nt}$ & $\kappa_t$ &  $\kappa_{nt}$ & $\kappa_t$ &  $\kappa_{nt}$ & $\kappa_t$ &  $\kappa_{nt}$ & $\kappa_t$ &  $\kappa_{nt}$ \\
\midrule
\multirow{2}{*}{$\mathcal{L}_{t} + \mathcal{L}_{nt} $} 
& CIFAR-10  & $78.8$ & $26.1$ & $84.6$ & $28.0$ & $84.0$ & $24.3$ & $84.2$ & $26.9$ & $82.9$ & $26.3$ \\
& ImageNet  & $74.0$ & $13.9$ & $94.0$ & $21.4$ & $82.0$ & $15.2$ & $72.0$ & $20.9$ & $80.5$ & $17.9$ \\
\midrule
\multirow{2}{*}{$\mathcal{L}^{\text{CE}}_{t} + \mathcal{L}_{nt} $} 
& CIFAR-10  & $77.4$ & $46.1$ & $89.4$ & $49.6$ & $88.8$ & $44.6$ & $88.1$ & $57.0$ & $85.9$ & $49.3$ \\
& ImageNet  & $66.0$ & $10.4$ & $98.0$ & $29.5$ & $92.0$ & $29.2$ & $78.0$ & $27.0$ & $83.5$ & $24.0$ \\
\midrule
\multirow{2}{*}{$\mathcal{L}_{t}$} 
& CIFAR-10  & $99.2$ & $97.9$ & $99.6$ & $98.4$ & $99.7$ & $99.3$ & $100$  & $99.3$ & $99.6$ & $98.7$ \\
& ImageNet  & $100$  & $68.5$ & $100$  & $76.7$ & $94.0$ & $67.3$ & $98.0$ & $85.4$ & $98.0$ & $74.5$ \\
\midrule
\multirow{2}{*}{$\mathcal{L}_{t1} + \mathcal{L}_{nt}$} 
& CIFAR-10  & $18.1$ & $3.4$  & $17.1$ & $2.7$  & $23.0$ & $5.8$  & $2.8$  & $4.9$  & $15.3$ & $4.2$  \\
& ImageNet  & $0.0$  & $0.0$  & $0.0$  & $0.1$  & $0.0$  & $0.0$  & $0.0$  & $0.1$  & $0.0$  & $0.1$  \\
\midrule
\multirow{2}{*}{$\mathcal{L}_{t2} + \mathcal{L}_{nt}$} 
& CIFAR-10  & $81.9$ & $32.4$ & $88.9$ & $38.9$ & $89.4$ & $34.4$ & $90.0$ & $35.3$ & $87.6$ & $35.3$ \\
& ImageNet  & $78.0$ & $23.7$ & $96.0$ & $31.4$ & $90.0$ & $22.7$ & $76.0$ & $30.1$ & $85.0$ & $27.0$ \\
\bottomrule
\end{tabular}
}
\end{table}

\begin{figure}[t]
    \centering
    \includegraphics[width=0.85\linewidth]{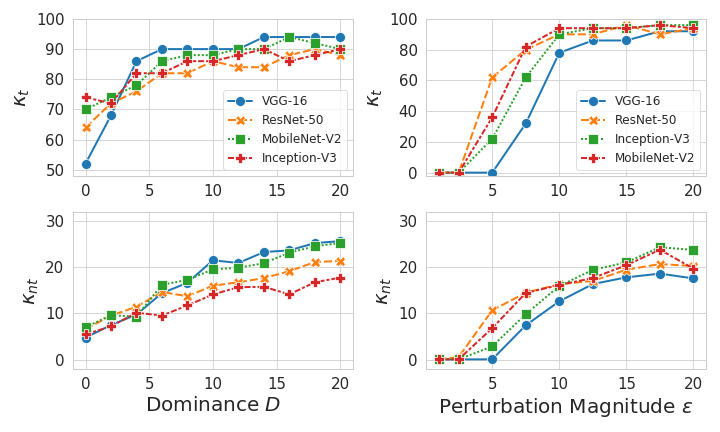}
    \caption{Analysis of the influence of the dominance value $D$ (left) and perturbation magnitude $\epsilon$ (right) on the targeted fooling ratios $\kappa_t$ (top) and $\kappa_{nt}$ (bottom) for the ImageNet dataset.}
    \label{fig:dominance_1}
\end{figure}

In the following, we perform ablation studies for the proposed DTA algorithm. All ablation experiments are performed on ResNet-20 for CIFAR-10 and ResNet-50 for the ImageNet dataset.

\subsubsection{Loss Function}
We perform an ablation study for the loss function design. In Table~\ref{tab:results_no_sink_loss} the performance of DTA for different loss function configurations is shown. We observe that our chosen loss design $\mathcal{L}_{t} +\mathcal{L}_{nt}$ achieves the best performance. In particular, we observe that excluding the non-targeted loss part $\mathcal{L}_{nt}$ results in a very high $\kappa_t$ close to $100\%$ for both, the CIFAR-10 and ImageNet dataset. However, the $\kappa_{nt}$ also increases drastically compared to the result obtained using $\mathcal{L}_{t} + \mathcal{L}_{nt}$. The average $\kappa_{nt}$ for ImageNet is $74.5\%$, and that for CIFAR-10 is even higher with a value of $98.7\%$. This clearly shows that under the absence of $\mathcal{L}_{nt}$, DTA fails to achieve the objective of being discriminative between samples from the targeted source class and non-targeted source classes. 
With the existence of $\mathcal{L}_{nt}$, the absence of either $\mathcal{L}_{t1}$ or $\mathcal{L}_{t2}$ also leads to inferior performance. Moreover, with the existence of $\mathcal{L}_{nt}$, we further explore another variant of $\mathcal{L}_{t}$ adopting the cross-entropy (CE) loss indicated as $\mathcal{L}^{\text{CE}}_{t}$. Similar to $\mathcal{L}_{t}$, $\mathcal{L}^{\text{CE}}_{t}$ is decomposed into two parts $\mathcal{L}^{\text{CE}}_{t1}$ and $\mathcal{L}^{\text{CE}}_{t2}$. $\mathcal{L}^{\text{CE}}_{t1}$ aims to reduce the logit value of the source class logit by calculating the negative cross-entropy between the network output and the one hot encoded source class label and $\mathcal{L}^{\text{CE}}_{t2}$ aims to increase the sink class logit by calculating the cross-entropy between the network output and the one hot encoded sink class label. 
We observe that this setup also achieves inferior performance compared to $\mathcal{L}_{t} +\mathcal{L}_{nt}$. The reason for this inferior performance can be attributed to the nature of the CE loss manipulating all logits, and not clamping the loss values. 

\subsubsection{Dominance Value $D$}
Further, we investigate the influence of the dominance value $D$ for clamping the loss part $\mathcal{L}_{t2}$. Fig.~\ref{fig:dominance_1} (left) shows the targeted fooling rates $\kappa_t$ and $\kappa_{nt}$ plotted over various dominance values. We observe that the value of $D$ has a significant influence on the behavior of the proposed DTA. Increasing $D$ increases both $\kappa_t$ and $\kappa_{nt}$. More specifically, $\kappa_t$ increases and saturates with further increasing $D$, while $\kappa_{nt}$ increases almost linearly with the increase of $D$. The results show that it is beneficial to choose an appropriate $D$ for achieving high $\kappa_t$ with relatively low $\kappa_{nt}$. However, here we only aim to show the influence of the hyper-parameter $D$ on the behavior of the proposed DTA and do not intend to find the optimal value which is dependent on the choice of models and dataset.

\subsubsection{Perturbation Magnitude $\epsilon$}
One constraint of adversarial perturbations is to be bound to a certain magnitude range. Here we investigate the influence of the perturbation magnitude $\epsilon$ and report the results in Fig~\ref{fig:dominance_1} (right). A sharp increase of $\kappa_t$ can be observed for $\epsilon$ values between $2.5$ and $10$ saturating around a targeted fooling ratio of $90\%$ for further increased $\epsilon$ values, while $\kappa_{nt}$ increases more steadily with increasing $\epsilon$ values.

\subsubsection{Weighting Factor $\alpha$}
One way an attacker can control the behavior of DTA is by manipulating the weighting factor $\alpha$ in Eq.~\ref{eq:loss}. In Table~\ref{tab:influence_alpha} we evaluate the influence of $\alpha$ on $\kappa_t$ and $\kappa_{nt}$. Higher values of $\alpha$ lead to lower values of $\kappa_{nt}$, since $\alpha$ weights the contribution of $\mathcal{L}_{nt}$ to the final loss value. Even though this behavior is desired, $\kappa_t$ decreases simultaneously.
For an effective attack, an attacker might consider a large gap between $\kappa_t$ and $\kappa_{nt}$, where neither a too large nor too small $\alpha$ is beneficial.

\begin{table}[t]
\centering
\caption{Influence of $\alpha$ in Eq.~\ref{eq:loss} on the targeted fooling ratios $\kappa_t$ and $\kappa_{nt}$.}
\label{tab:influence_alpha}
\small
\setlength\tabcolsep{2.0pt}
\scalebox{0.95}{
\begin{tabular}{c |cc|cc|cc|cc|cc}
\toprule
\multirow{2}{*}{Dataset} & \multicolumn{2}{c}{$0.1$} & \multicolumn{2}{c}{$0.5$} & \multicolumn{2}{c}{$1$} & \multicolumn{2}{c}{$2$} & \multicolumn{2}{c}{$10$} \\
& $\kappa_t$ &  $\kappa_{nt}$ & $\kappa_t$ &  $\kappa_{nt}$ & $\kappa_t$ &  $\kappa_{nt}$ & $\kappa_t$ &  $\kappa_{nt}$ & $\kappa_t$ &  $\kappa_{nt}$ \\
\midrule
CIFAR-10  & $98.0$ & $68.4$ & $90.1$ & $37.8$ & $84.6$ & $28.0$ & $70.8$ & $15.3$ & $29.4$ & $ 5.1$ \\
ImageNet  & $98.0$ & $60.6$ & $94.0$ & $31.5$ & $94.0$ & $21.4$ & $90.0$ & $9.0$ & $0.0$   & $0.1$  \\
\bottomrule
\end{tabular}
}
\end{table}

\begin{table}[t]
\centering
\caption{Influence of number of training samples per class on the targeted fooling ratios $\kappa_t$ and $\kappa_{nt}$.}
\label{tab:samples_per_class}
\small
\setlength\tabcolsep{2.0pt}
\scalebox{0.95}{
\begin{tabular}{c |cc|cc|cc|cc|cc}
\toprule
\multirow{2}{*}{Dataset} & \multicolumn{2}{c}{$50$} & \multicolumn{2}{c}{$100$} & \multicolumn{2}{c}{$250$} & \multicolumn{2}{c}{$500$}  & \multicolumn{2}{c}{$1000$} \\
& $\kappa_t$ &  $\kappa_{nt}$ & $\kappa_t$ &  $\kappa_{nt}$ & $\kappa_t$ &  $\kappa_{nt}$ & $\kappa_t$ &  $\kappa_{nt}$ & $\kappa_t$ &  $\kappa_{nt}$ \\
\midrule
CIFAR-10  & $46.7$ & $18.4$ & $60.2$ & $20.3$ & $73.9$ & $22.7$ & $80.9$ & $27.6$ & $83.2$ & $27.8$ \\
ImageNet  & $50.0$ & $2.9$  & $64.0$ & $4.3$  & $86.0$ & $12.5$ & $94.0$ & $19.0$ & $96.0$ & $17.7$ \\
\bottomrule
\end{tabular}
}
\end{table}

\subsubsection{Number of Available Training Samples}
Finally, we investigate the influence of the available number of training samples on the attack behavior. In Table~\ref{tab:samples_per_class} we report the influence of the number of available training samples per class on the attack performance. With the same number of training iterations, we find that a smaller number of training samples per class lead to lower $\kappa_t$ and $\kappa_{nt}$ and the gap between $\kappa_t$ and $\kappa_{nt}$ decreases accordingly. However, with as small as $50$ samples per class, the algorithm still works reasonably well. For example, for ImageNet $\kappa_t$ is $50\%$ while $\kappa_{nt}$ is as low as $2.9\%$.
\begin{table}[!htbp]
\centering
\caption{Quantitative results for the generated DT-Patch on ImageNet}
\label{tab:dta-result}
\small
\setlength\tabcolsep{2.0pt}
\scalebox{0.95}{
\begin{tabular}{cc|cc|cc}
\toprule
\multicolumn{2}{c}{Hammer $\rightarrow$ Hummingbird} & \multicolumn{2}{c}{Screwdriver $\rightarrow$ Go-Kart} & \multicolumn{2}{c}{Coffee Mug $\rightarrow$ Chocolate Sauce} \\
$\kappa_t$ &  $\kappa_{nt}$ & $\kappa_t$ &  $\kappa_{nt}$ & $\kappa_t$ &  $\kappa_{nt}$ \\
\midrule
$80.0$ & $42.7$ & $92.0$ & $44.9$ & $96.0$ & $41.6$ \\
\bottomrule
\end{tabular}
}
\end{table}
\section{Double Targeted Patch}
\begin{figure}[!htbp]
    \centering
    \includegraphics[width=0.2\linewidth]{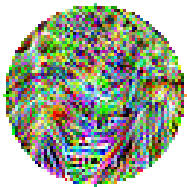}
    \includegraphics[width=0.25\linewidth]{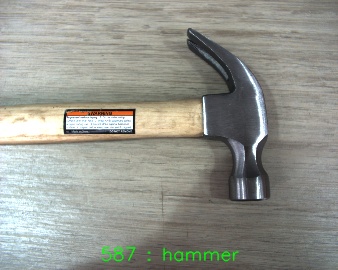}
    \includegraphics[width=0.25\linewidth]{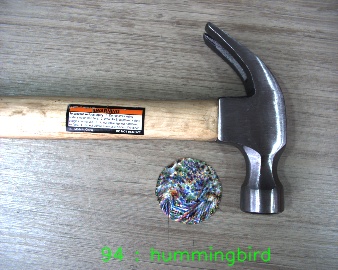}
    \includegraphics[width=0.25\linewidth]{imgs/patch1_hammer_hummingbird.jpg}
    
    \includegraphics[width=0.2\linewidth]{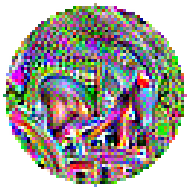}
    \includegraphics[width=0.25\linewidth]{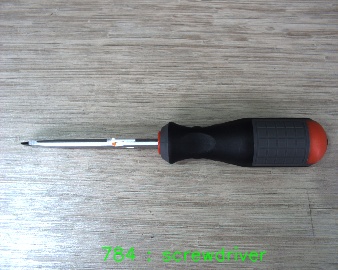}
    \includegraphics[width=0.25\linewidth]{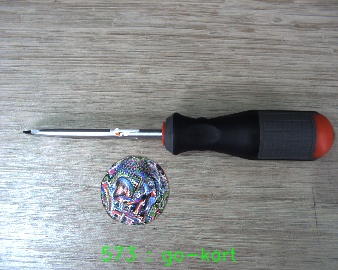}
    \includegraphics[width=0.25\linewidth]{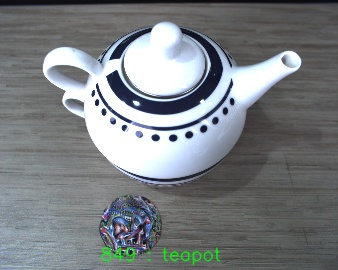}
    
    \includegraphics[width=0.2\linewidth]{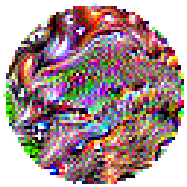}
    \includegraphics[width=0.25\linewidth]{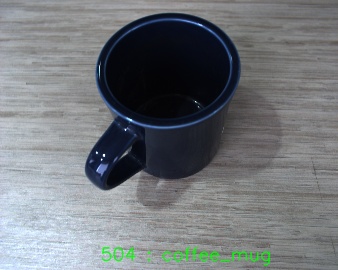}
    \includegraphics[width=0.25\linewidth]{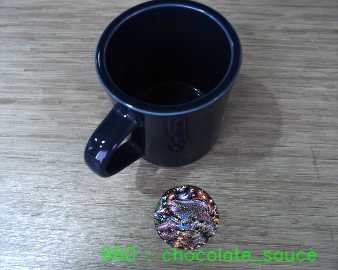}
    \includegraphics[width=0.25\linewidth]{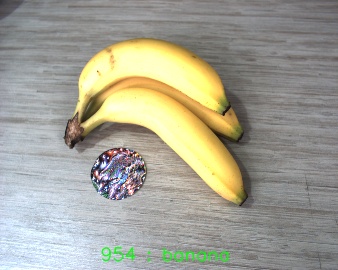}
    \caption{Real-world examples of the DT-Patch for three different scenarios (see Table~\ref{tab:dta-result}).}
    \label{fig:physical_patch}
\end{figure}
We extend DT-UAP to a physical-world attack~\cite{brown2017adversarial,liubias2020bias} by generating a physical patch. We apply the concept of the DTA to attack one source class to a sink class. We choose VGG-16 trained on ImageNet as the target network. For generating a physical patch, we restrict the perturbation to a circular area, as well as its magnitude to lie in image range, i.e.\ $x+\delta \in [0,1]$. We show three cases by choosing the source-sink class pairs as indicated in Table~\ref{tab:dta-result}. 
Despite being a more challenging scenario than the original adversarial patch, we observe from Table~\ref{tab:dta-result} that $\kappa_t$ is larger than $\kappa_{nt}$ by a non-trivial margin. This indicates that the patch fulfills the objective. The qualitative results in Figure~\ref{fig:physical_patch} show the applied patch fooling the source into the sink class, while having no influence on a sample from a non-targeted class. 

\section{Conclusion and Future Work}
We proposed DTA to extend the exisitng UAP and CD-UAP for a more flexible attack control. The generated DT-UAP shifts one predefined source class into one predefined sink class, simultaneously attempts to minimize the targeted fooling ratio for samples from the non-targeted source classes. The effectiveness of DTA is demonstrated with extensive experiments on multiple datasets for different network architectures. We further presented an extension of DTA to the Multi2One scenario, driving multiple source classes into one sink class. With some preliminary results we found it also worked for a very challenging Multi2Multi scenario with limited success, and leave further explorations for future work.

\bibliographystyle{splncs}
\bibliography{bibliography}

\end{document}